\documentclass[sigconf]{acmart}

\usepackage[normalem]{ulem}
\usepackage[final]{pdfpages}
\usepackage{balance}
\usepackage{balance} 

\AtBeginDocument{%
  \providecommand\BibTeX{{%
    \normalfont B\kern-0.5em{\scshape i\kern-0.25em b}\kern-0.8em\TeX}}}

\copyrightyear{2023} 
\acmYear{2023} 
\setcopyright{acmlicensed}\acmConference[SIGIR '23]{Proceedings of the 46th International ACM SIGIR Conference on Research and Development in Information Retrieval}{July 23--27, 2023}{Taipei, Taiwan}
\acmBooktitle{Proceedings of the 46th International ACM SIGIR Conference on Research and Development in Information Retrieval (SIGIR '23), July 23--27, 2023, Taipei, Taiwan}
\acmPrice{15.00}
\acmDOI{10.1145/3539618.3591975}
\acmISBN{978-1-4503-9408-6/23/07}

\begin{document}

\definecolor{ds}{RGB}{127,201,127}
\definecolor{zm}{RGB}{255,0,0}
\definecolor{jgs}{RGB}{253,192,134}

\newcommand{\cl}[1]{{\bf \color{zm} [CL: #1]}}
\newcommand{\ds}[1]{{\bf \color{ds} [DW: #1]}}
\newcommand{\jgs}[1]{{\bf \color{jgs} [JGS: #1]}}


\title{DocGraphLM: Documental Graph Language Model for Information Extraction}

\author{Dongsheng Wang}
\affiliation{%
  \institution{JPMorgan AI Research}
  \streetaddress{25 Bank St}
  \city{London}
  \country{UK}}
\email{dongsheng.wang@jpmchase.com}

\author{Zhiqiang Ma}
\affiliation{%
  \institution{JPMorgan AI Research}
  \streetaddress{383 Madison Ave}
  \city{New York}
  \state{New York}
  \country{USA}}
\email{zhiqiang.ma@jpmchase.com}

\author{Armineh Nourbakhsh}
\affiliation{%
  \institution{JPMorgan AI Research}
  \streetaddress{383 Madison Ave}
  \state{New York}
  \city{New York}
  \country{USA}}
\email{armineh.nourbakhsh@jpmchase.com}

\author{Kang Gu}
\affiliation{%
  \institution{Dartmouth College}
  \city{Hanover}
  \state{New Hampshire}
  \country{USA}}
\email{Kang.Gu.GR@dartmouth.edu}

\author{Sameena Shah}
\affiliation{%
  \institution{JPMorgan AI Research}
  \streetaddress{383 Madison Ave}
  \city{New York}
  \state{New York}
  \country{USA}}
\email{sameena.shah@jpmchase.com}






\begin{abstract}
Advances in Visually Rich Document Understanding (VrDU) have enabled information extraction and question answering over documents with complex layouts. Two tropes of architectures have emerged---transformer-based models inspired by LLMs, and Graph Neural Networks. In this paper, we introduce DocGraphLM, a novel framework that combines pre-trained language models with graph semantics. To achieve this, we propose 1) a joint encoder architecture to represent documents, and 2) a novel link prediction approach to reconstruct document graphs. DocGraphLM predicts both directions and distances between nodes using a convergent joint loss function that prioritizes neighborhood restoration and downweighs distant node detection. Our experiments on three SotA datasets show consistent improvement on IE and QA tasks with the adoption of graph features. Moreover, we report that adopting the graph features accelerates convergence in the learning process druing training, despite being solely constructed through link prediction. 
\end{abstract}



\maketitle

\section{Introduction}

\begin{figure*}
\small
 \centering
\includegraphics[width=0.98\textwidth, trim ={1.2cm 5.3cm 1cm 7.6cm }, clip]{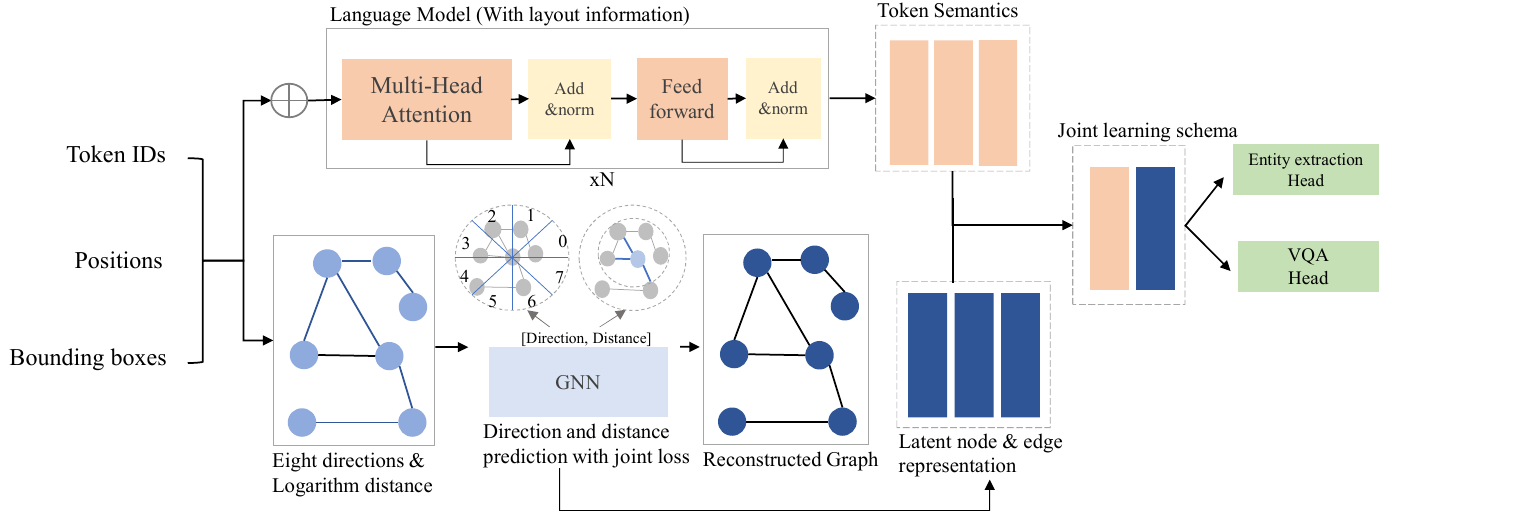}
  \caption{\label{fig:overallarchi}The model architecture of DocGraphLM.}
  \label{fig:teaser}
  \centering
\end{figure*}

Information extraction from visually-rich documents (VrDs), such as business forms, receipts, and invoices in the format of PDF or image  has gained recent traction. Tasks such as field identification and extraction and entity linkage are crucial to digitizing VrDs and building information retrieval systems on the data. Tasks that require complex reasoning such as Visual Question Answering over documents require modeling the spatial, visual, and semantic signals in VrDs. Therefore, VrD Understanding is concerned with modeling the multi-modal content in image documents. Previous research has explored the use of encoding text, layout, and image features in a layout language model or multi-modal setting to improve downstream tasks. For example, LayoutLM and its variants \cite{xu2020layoutlm, xu2020layoutlmv2, huang2022layoutlmv3} use image and layout information to enhance the representation of text, thereby improving performance on various tasks. However, models using Transformer mechanisms pose a challenge to representing spatially distant semantics, such as table cells far from their headers or contents across line breaks. In light of these limitations, a few studies \cite{yao2019graph,zhang2020every} have proposed using graph neural networks (GNNs) to model relationships and structures between text tokens or segments in documents. Although these models alone still underperform layout language models, they demonstrate the potential of incorporating additional structured information to improve document representation.

Motivated by this, we introduce a novel framework called DocGraphLM that integrates document graph semantics and the semantics derived from pre-trained language models to improve document representation. As depicted in Figure \ref{fig:overallarchi}, the input to our model is embeddings of tokens, positions, and bounding boxes, which form the foundation of the document representation. To reconstruct the document graph, we propose a novel link prediction approach that predicts directions and distances between nodes by using a joint loss function, which balances the classification and regression loss. Additionally, the loss encourages close neighborhood restoration while downgrading detections on farther nodes. This is achieved by normalizing the distance through logarithmic transformation, treating nodes separated by a specific order-of-magnitude distance as semantically equidistant.

Our experiments on multiple datasets including FUNSD, CORD, and DocVQA, show the superiority of the model in a consistent manner. Furthermore, the incorporation of graph features is found to accelerate the learning process. We highlight the main contributions of our work as follows:
\begin{itemize}
\item  we propose a novel architecture that integrates a graph neural network with pre-trained language model to enhance document representation;
\item  we introduce a link prediction approach to document graph reconstruction, and a joint loss function that emphasizes restoration on nearby neighbor nodes;
\item  lastly, the proposed graph neural features result in a consistent improvement in performance and faster convergence.
\end{itemize}

\section{Related Work}
Transformer-based architectures have been successfully applied to layout understanding tasks, surpassing previous state-of-the-art (SotA) results \cite{wang2020docstruct,majumder2020representation,wang2021layoutreader,li2021selfdoc,Lukasz2020,li2021structext}. Studies such as LayoutLM \cite{xu2020layoutlm} and LayoutLMv2 \cite{xu2020layoutlmv2} fuse text embeddings with visual features using a region proposal network, allowing the models to be trained on objectives such as Masked Visual Language Model (MVLM) and spatial aware attention, resulting in improved performance on complex tasks such as VQA and form understanding. TILT \cite{l2021going} augments the attention by adding bias to capture relative 2-D positions, which has shown excellent performance on DocVQA leaderboard. StructuralLM \cite{li2021structurallm} makes the most of the interactions of cells where each cell shares the same bounding boxes. 


The use of GNNs \cite{scarselli2008graph} to represent documents allows information to propagate more flexibly. In GNN-based VrDU models, documents are often represented as graphs of tokens and/or sentences, and edges represent spatial relationships among them, e.g. capturing K-Nearest Neighbours. GNN-based models can be used for various document-grounded tasks such as text classification \cite{yao2019graph,zhang2020every} or key information extraction \cite{Brian2021, yu2021pick}. 
However, their performance still lags behind that of layout language models. This is because graph representation alone is insufficient to capture the rich semantics of a document. In cases where GNN-based models substantially outperform layout language models, they are often larger and focused on specific tasks \cite{lee2022formnet}.
In this paper, we propose a framework that combines the rich semantics of layout language models with the robust structural signal captured by GNN models. We demonstrate how the addition of graph semantics can enhance the performance of layout language models on IE and QA tasks, and improve model convergence. 

\section{DocGraphLM: Document Graph Language Model}
\subsection{Representing document as graph}
In GNN, a graph consists of nodes and edges. In the context of representing document as graph, the nodes represent text segments (i.e. groups of adjacent words) and the relationships between them are represented as edges. Text segments from image documents can be obtained through Optical Character Recognition tools, which often capture the tokens as bounding boxes of various sizes.

To generate the edges between nodes, we adopt a novel heuristic named Direction Line-of-sight (D-LoS), instead of the commonly used K-nearest-neighbours (KNN) \cite{graphie2019} or $\beta$-skeleton approach \cite{rope2021}. The KNN approach may result in dense, irrelevant rows or columns being treated as neighbours, ignoring the fact that some key-value pairs in a form can be farther apart nodes. To address this, we adopt the D-LoS approach, where we divide the 360-degree horizon surrounding a source node into eight discrete 45-degree sectors, and we determine the nearest node with respect to the source node within each sector. These eight sectors define eight directions with respect to the source node. This definition is inspired by the pre-training task reported in StrucTexT \cite{li2021structext} which applies this approach to construct its graph representation.

\paragraph{Node representation.}A node has two features --- text semantics and node size. The text semantics can be obtained through token embeddings (e.g. from language models), while the node size is expressed by its dimensions on $x$ and $y$ coordinates, mathematically $M =\text{emb}([width, height])$ were $width = x_2-x_1$ and $ height = y_2 - y_1$, given that $(x_1, y_1)$ and $(x_2, y_2)$ are the coordinates of top left corner and bottom right corner of the segment bounding box. Intuitively, the node size is a significant indicator because it helps differentiate font size and potentially the semantic role of the segment, e.g., title, caption, and body. Thus, we denote a node input as $E_{u} = \text{emb}(T_u) \oplus M_u$, where $u = \{1, 2, ..., N\}$ indicates the $u$th node in a document and $T_u$ stands for the texts inside the node $u$.

We learn the node representation by reconstructing the document graph using GNN, expressed as $h_{u}^{G}=\text{GNN}(E_{u})$. Details on learning $h_{u}^{G}$ are described in Section~\ref{sec:reconstructing}.

\paragraph{Edge representation.} To express the relationships between two nodes, we use their polar features, including relative distance and direction (one of eight possibilities). We compute the shortest Euclidean distance, $d$, between the two bounding boxes. To reduce the impact of distant nodes that may be less semantically relevant to the source node, we apply a distance smoothing technique with log transformation denoted as $e_{\text{dis}}=\log(d+1)$. The relative direction  $e_{\text{dir}}\in\{0,\ldots,7\}$ for a pair of nodes is obtained from D-LoS. We define a linkage, denoted as $e_{p} = [e_{\text{dis}}, e_{\text{dir}}]$, to reconstruct the document graph in section \ref{sec:reconstructing}.



\subsection{Reconstructing graph by link prediction}
\label{sec:reconstructing}


We predict two key attributes of the linkages $e_{p}$ to reconstruct the graph and frame the process as a multi-task learning problem.

The input to the GNN is the encoded node representations, and the representation is passed through the message passing mechanism on GNN, specifically:
\begin{equation}
\small
h_u^{G, l+1} := \text{aggregate}({h_v^{G,l}, \forall v \in \mathcal{N}(u)}),
\end{equation}
where $l$ is the layer of neighbors, $\mathcal{N}(u)$ denotes the set of neighbors of node $u$, and $\text{aggregate}(\cdot)$ is an aggregation function that updates the node representation.

We jointly train the GNN on two tasks --- predicting the distance and direction between nodes --- to learn the node representation. For distance prediction, we define a regression head $\hat{y}^e_{u,v}$, which generates a scalar value through the dot-product of two node vectors, and uses a linear activation, as presented in Equation \ref{eq:linear_reg}.
\begin{equation}
\small
\label{eq:linear_reg}
    \hat{y}^e_{u,v} = Linear((h_u^{G })^{\top} \times h_v^{G}) 
\end{equation}

For direction prediction,  we define a classification head $\hat{y}^{d}_{u,v}$ that assigns one of eight directions to each edge based on the element-wise product between two nodes, expressed as follows:
\begin{equation}
\small
    \hat{y}^{d}_{u,v} = \sigma((h_u^{G} \odot h_v^{G}) \times W)
\end{equation}
where $h_u^{G} \odot h_v^{G}$ is an element-wise product between two nodes and $W$ is the learnable weight for the product vector. $\sigma$ is a non-linear activation function.

We use MSE loss for distance regression and cross-entropy for the direction classification, respectively. Then, the joint loss is:
\begin{equation}
\small
\begin{aligned}
    loss = \sum_{(u,v)\in \text{batch}}[(\lambda \cdot \text{loss}^{\text{MSE}}(\hat{y}^{e}_{u,v}, y^{e}_{u,v}) \\ 
    + (1 - \lambda) \cdot \text{loss}^{\text{CE}}(\hat{y}^{d}_{u,v}, y^{d}_{u,v})] \cdot (1 - r_{u,v})
\end{aligned}
\end{equation}
where $\lambda$ is a tunable hyper-parameter that balances the weights of the two losses, and $r_{u,v}$ is the normalization of the distance $e_{\text{dis}}$, constrained to the interval $[0,1]$, so that the value of $1-r_{u,v}$ downweights distant segments and favors nearby segments.

\subsection{Joint representation}
The joint node representation, $h_{u}^C$, is a combination of the language model representation $h_{u}^{L}$ and the GNN representation $h_{u}^{G}$ through an aggregation function $f$ (e.g., concatenation, mean, or sum) represented as $h_{u}^C = f(h_{u}^{L}, h_{u}^{G})$. In this work, we operationalize the aggregation function $f$ with concatenation at the token level. The introduced node representations can be utilized as input for other models to facilitate downstream tasks, e.g., $\text{IE\_Head}(h_u^{C})$ for entity extraction and $\text{QA\_Head}(h_u^{C})$ for visual question answering task.


\section{Experiments}
\subsection{Datasets and baselines}
\label{sc:dataset}
We evaluate our models on two information extraction tasks across three commonly used datasets: FUNSD \cite{jaume2019funsd}, CORD \cite{park2019cord}, and DocVQA \cite{mathew2021docvqa}. FUNSD and CORD focus on entity-level extraction, while DocVQA concentrates on identifying answer spans in image documents in a question-answering task. 
Dataset statistics are shown in Table \ref{tb:statis_dataset}. Please refer to the citations for more details.

It is noted that the OCR files provided in DocVQA\footnote{https://www.docvqa.org/} contain a small number of imperfect OCR outputs, e.g., text misalignment and missing texts, which leads to failures in identifying the answers. We can only use 32,553 samples for training and 4,400 samples for validation. We denote the modified dataset as $DocVQA^\dagger$. In the interest of ensuring fair comparison in our experiments, we have maintained the use of the OCR outputs from the dataset.

As our baselines, we employ the SotA models that make use of different features, including RoBERTa \cite{roberta2019}, BROS \cite{hong2020bros}, DocFormer-base \cite{appalaraju2021docformer}, StructuralLM \cite{li2021structurallm}, LayoutLM \cite{xu2020layoutlm}, LayoutLMv3 \cite{huang2022layoutlmv3} and Doc2Graph \cite{gemelli2022doc2graph}. RoBERTa is transformer model without any layout or image features, BROS and StructuralLM adopt layout information solely, DocFormer and LayoutLMv3 utilizes both layout and image features, and Doc2Graph soly relies on document graph features.

\begin{table}[]
\small
\caption{\label{tb:statis_dataset}Statistics of visual document datasets. The differences between DocVQA and DocVQA$^\dagger$ is introduced in Section \ref{sc:dataset}.}
\begin{tabular}{@{}lllll@{}}
\toprule
Dataset & No. labels & No. train & No. val & No. test \\ \midrule
FUNSD   & 4             & 149          & -            & 50          \\
CORD    & 30            & 800          & 100          & 100         \\
DocVQA  & -             & 39,000       & 5,000        & 5,000       \\
DocVQA$^\dagger$   & - & 32,553           & 4,400          & 5,000          \\     \bottomrule
\end{tabular}
\end{table}

\subsection{Experimental setup}
For FUNSD and CORD, we adopt the following training hyper-parameters: epoch = 20, learning rate = 5e-5, and batch size = 6, and trained our model on a single NVIDIA T4 Tensor Core GPU. For DocVQA, we apply the following training hyper-parameters: epoch = 5, learning rate = 5e-5, and batch size = 4.

We adopt GraphSage \cite{hamilton2017inductive} as our GNN model, as it has been proven effective in document graph features\cite{gemelli2022doc2graph}.
For graph reconstruction, we set a constant value $\lambda$=0.5 throughout the experiment. 

\subsection{Results}
The performance of DocGraphLM and other models on the FUNSD dataset are presented in Table~\ref{tb:funsdres}. Our model reaches the best F1 score at 88.77, achieved when it is paired with the LayoutLMv3-base model. On the other hand, RoBERTa-base (which does not leverage layout features) has the lowest F1 score of 65.37, but combining it with DocGraphLM results in a 1.66 point improvement. Please note scores with $^\diamond$ are reported in the corresponding citations. The same notation applies to other tables.

For the CORD dataset, the performance comparisons are shown in Table \ref{tb:cordres}, and the best performance is achieved by DocGraphLM (LayoutLMv3-base) with an F1 score of 96.93, followed closely by BROS. Similarly, even though RoBERTa-base alone achieves a much lower score, DocGraphLM (RoBERTa-base) increases the F1 score by 2.26 points.

Table \ref{tb:dcvqares} shows the model performance on the DocVQA test dataset. The performance scores are obtained by submitting our model output to the DocVQA leaderboard\footnote{\url{https://rrc.cvc.uab.es/?ch=17\&com=evaluation&task=1}}, as ground-truth answers are not provided to the public. Besides the overall score, the model's performances on sub-category tasks are also reported. DocGraphLM (with LayoutLMv3-base) outperforms others in almost every aspect except pure text semantics, which shows the model's ability to model multi-modal semnatics effectively. 
The table presents strong evidence towards the efficiency of DocGraphLM in improving document representations, when layout language models are augmented with our approach.


The superior performance across various datasets indicates that using the graph representation proposed in DocGraphLM leads to consistent improvements. A p-value less than 0.05 was received when comparing the models' performance across these datasets, indicating a statistically significant improvement from our model.

 

%
\begin{table}[]
\small
\caption{\label{tb:funsdres}Model performance comparison on FUNSD.}
\resizebox{\columnwidth}{!}{%
\begin{tabular}{@{}lllll@{}}
\toprule
Model               & F1 & Precision & Recall &  \\ \midrule
RoBERTa-base         &  65.37           &   61.17          &  70.20    & \\
Doc2Graph$^\diamond$\cite{gemelli2022doc2graph}      &    82.25      &      -     &  -      &  \\
StructuralLM\_large$^\diamond$\cite{li2021structurallm}   &    85.14      &      83.52    &  86.81      &  \\
LayoutLM-base$^\diamond$\cite{xu2020layoutlm}       &     78.66     &   75.97        &     81.55   &  \\ 
LayoutLMv3-base\cite{huang2022layoutlmv3}         &   88.16      &     86.70    &  87.7      &  \\ 
BROS$^\diamond$\cite{hong2020bros}   &   83.05       &   81.16     &    85.02   &  \\ 
DocFormer-base$^\diamond$\cite{appalaraju2021docformer}  & 83.34 &  80.76 & 86.09 &  \\
\midrule
DocGraphLM (RoBERTa-base)      &    67.03 ($\uparrow$1.66)     &    62.92       &   70.0     &   \\
DocGraphLM (LLMv3-base) &     \textbf{88.77($\uparrow$0.61)}     &      \textbf{87.44}      &   \textbf{90.15}      &  \\ \bottomrule
\end{tabular}
}
\end{table}

\begin{table}[]
\small
\caption{\label{tb:cordres} Model performance comparison CORD.}
\resizebox{\columnwidth}{!}{%
\begin{tabular}{@{}lllll@{}}
\toprule
Model               & F1 & Precision & Recall &  \\ \midrule
RoBERTa-base             &   48.99        &     42.77      &  57.34   & \\
LayoutLM-base$^\diamond$      &     94.80    &   95.03        &   94.58     &  \\ 
LayoutLMv3-base   &  95.59      &    95.31    &   95.88        &  \\ 
BROS$^\diamond$       &   95.36     &     95.58      &   95.14     &  \\ 
DocFormer-base$^\diamond$ & 96.33 & 96.52 & 96.14 & \\ 
\midrule
DocGraphLM (RoBERTa-base)      &   51.25 ($\uparrow$2.26)    &    45.45     &  58.76   &   \\
DocGraphLM (LayoutLMv3-base) &   \textbf{96.93}  ($\uparrow$1.62)   &     \textbf{96.86}       &    \textbf{97.01}     &  \\ \bottomrule
\end{tabular}
}
\end{table}

\begin{table}[]
\small	
\caption{\label{tb:dcvqares} Model performance comparison on DocVQA testing dataset. Scores are from DocVQA leaderboard.}

\resizebox{\columnwidth}{!}{%
\begin{tabular}{@{}llllll@{}}
\toprule
Model               & Score & Form &  Table  & Text\\ \midrule
RoBERTa\_base     & 60.40     &  71.75      &  54.23   &  61.35 & \\
LayoutLMv3\_base          &   67.80       &   77.84        &    67.58   & \textbf{70.55} &  \\ \midrule
DocGraphLM (LayoutLMv3-base) &  \textbf{69.84} ($\uparrow$2.04)       &  \textbf{79.73}     & \textbf{68.48}     & 63.23  &  \\ \bottomrule
\end{tabular}
}
\end{table}

\subsection{Impact on convergence}

\begin{figure}[htp]
    \footnotesize	
    \centering
    \includegraphics[width=6cm]{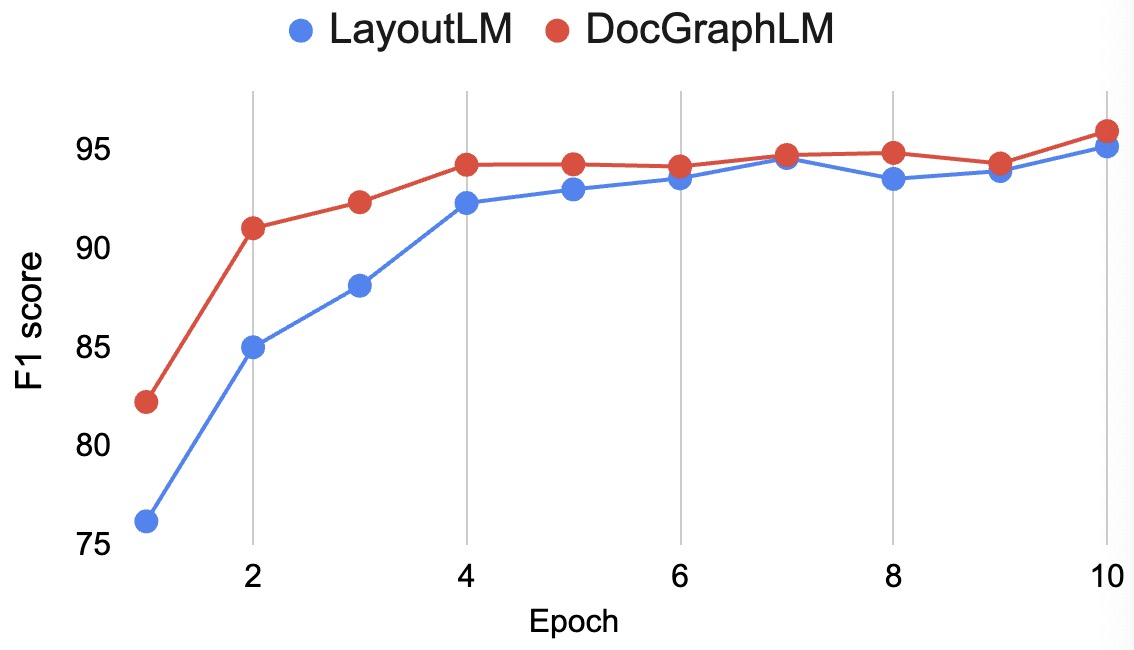}
    \caption{\label{fig:convergspeed}Model convergence speed comparison on  CORD. The curves are generated from averaging over ten trials.}
    \label{fig:galaxy}
\end{figure}

We also observed that the training convergence speed is often faster when supplementing the graph features than vanilla LayoutLM (V1 and V3 base models). For example, Figure~\ref{fig:convergspeed} illustrates that the F1 score improves in a faster convergence rate within the first four epochs, when testing on the CORD dataset. This could be due to the graph features allowing the transformer to focus more on the nearby neighbours, which eventually results in a more effective information propagation process. 

\section{Conclusion and Future Work}
 This paper presents a novel DocGraphLM framework incorporating graph semantics with pre-trained language models to improve document representation for VrDs. The proposed linkage prediction method reconstructs the distance and direction between nodes, increasingly down-weighting more distant linkages. Our experiments on multiple downstream tasks on various datasets show enhanced performance over LM-only baseline. Additionally, introducing the graph features accelerates the learning process.
As a future direction, we plan to incorporate different pre-training techniques for different document segments. We will also examine the effect of different linkage representations for graph reconstruction.



\paragraph{Disclaimer}
\footnotesize
This paper was prepared for informational purposes by the Artificial
Intelligence Research group of JPMorgan Chase \& Co. and its affiliates (“JP Morgan”),
and is not a product of the Research Department of JP Morgan. JP Morgan makes no
representation and warranty whatsoever and disclaims all liability, for the completeness,
accuracy or reliability of the information contained herein. This document is
not intended as investment research or investment advice, or a recommendation, offer
or solicitation for the purchase or sale of any security, financial instrument, financial
product or service, or to be used in any way for evaluating the merits of participating
in any transaction, and shall not constitute a solicitation under any jurisdiction or to
any person, if such solicitation under such jurisdiction or to such person would be
unlawful.

\bibliographystyle{ACM-Reference-Format}
\balance
\bibliography{references}


\end{document}